\documentclass[11pt,letterpaper]{article}

\usepackage[T1]{fontenc}
\usepackage{newtxtext,newtxmath}
\usepackage[letterpaper,margin=1in]{geometry}
\usepackage{microtype}
\usepackage{graphicx}
\usepackage{booktabs}
\usepackage{amsfonts}
\usepackage{amsmath}
\usepackage{array}
\usepackage{natbib}
\usepackage{caption}
\usepackage{placeins}
\usepackage[hyphens]{url}
\usepackage[hidelinks]{hyperref}

\hypersetup{
    pdftitle={V-Engram: Trigger-Indexed External Memory for Modular Text-to-Image Personalization},
    pdfauthor={Haoran He, Runyuan Cai, Yiming Wang, Yu Lin, Xiaodong Zeng}
}

\title{V-Engram: Trigger-Indexed External Memory for Modular Text-to-Image Personalization}
\author{Haoran He \quad Runyuan Cai \quad Yiming Wang\\
Yu Lin \quad Xiaodong Zeng\\[0.4em]
\textit{AutoArk-AI}}
\date{}

\newcommand{\method}{V-Engram}
\newcommand{\sdthree}{Stable Diffusion 3.5}
\newcommand{\mmdit}{MMDiT}
\newcommand{\resultfigurewidth}{0.74\textwidth}
\newcommand{\sampletrigger}{\texttt{\textless dog\textgreater}}

\begin{document}

\maketitle

\begin{center}
    \centering
    \includegraphics[width=\textwidth]{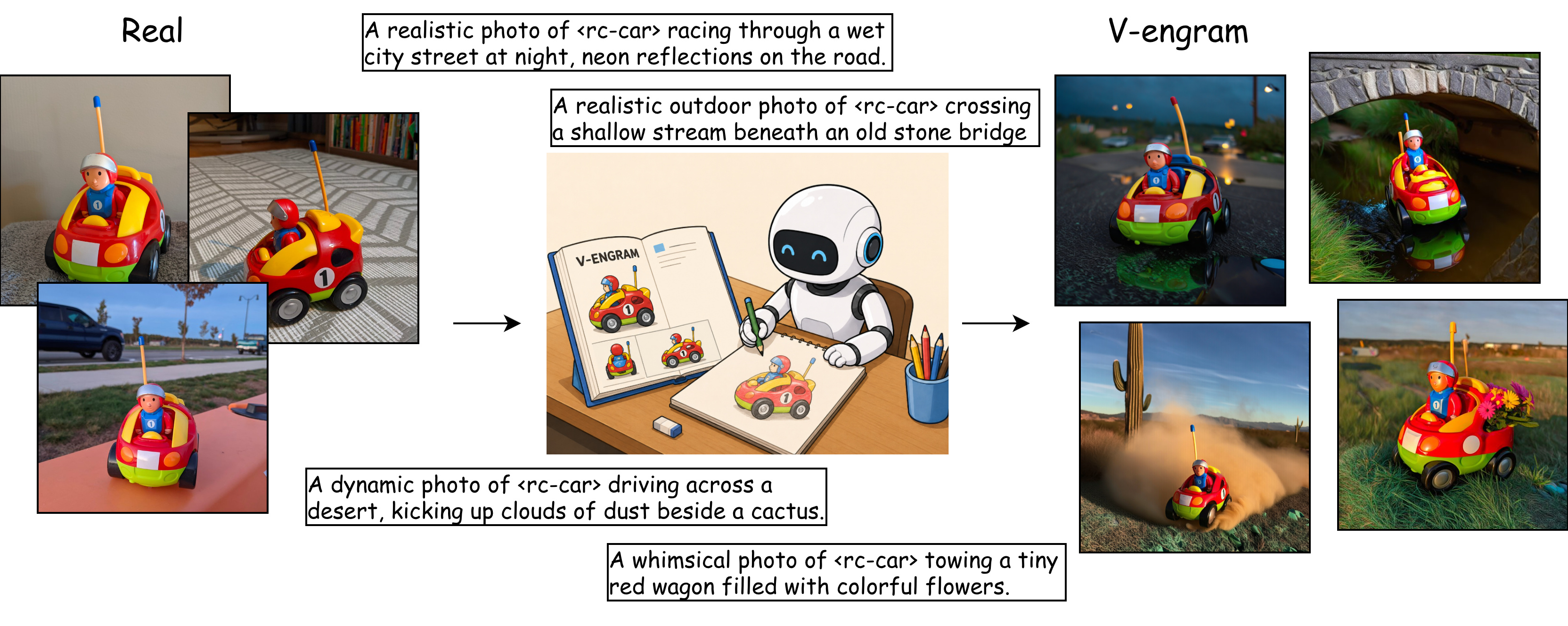}
    \captionsetup{type=figure,hypcap=false}
    \caption{Compositional generation with \method{}. A few reference images bind the trigger \texttt{\textless rc-car\textgreater} to a specific toy car; three are shown for illustration. The same memory composes with unseen environments, motion, lighting, and object interactions while preserving the characteristic driver, antenna, colors, and vehicle shape.}
    \label{fig:front-showcase}
\end{center}

\begin{abstract}
Pretrained text-to-image models contain broad visual knowledge, yet they cannot reliably acquire or refine a specific visual identity from only a few references while preserving compositional control. Token-embedding methods are compact but often underfit identity, whereas adapter-based methods improve fidelity through persistent weight updates that can be costly to store and interfere when concepts are composed. We introduce \method{}, a trigger-indexed external memory mechanism for \sdthree{}. Each concept is assigned an explicit trigger that retrieves concept-specific memory, whose gated directions enter frozen text-encoder and \mmdit{} context states as relative residuals. Separating this memory from backbone adaptation enables prompt-selective and multi-concept access without merging model updates. Experiments show that \method{} broadly matches DreamBooth-LoRA in overall subject fidelity while showing advantages in settings such as contextual subject preservation. Prompt-matched loading retrieves only matched entries, reducing most additional adaptation-state loading for a single-concept query. Qualitative results further demonstrate paired-trigger composition and same-class separation, while prompts without registered entries retain the frozen model's base behavior. Together, these results establish trigger-indexed memory as a modular interface for adding targeted visual evidence without rewriting the generator.
\end{abstract}

\section{Introduction}

Large text-to-image generators encode broad visual and compositional knowledge \citep{rombach2022latent,esser2024scaling}, yet their coverage is neither complete nor uniformly precise. A model may miss a newly observed subject, confuse similar instances, or render a familiar identity only approximately. Personalization thus spans both acquiring new concepts and selectively refining imprecise prior knowledge. This raises a broader question: where should targeted visual evidence be stored, how should a prompt retrieve it, and how can it enter generation without overwriting useful priors?

Few-shot personalization is a natural testbed for this question \citep{gal2023textual,ruiz2023dreambooth,wu2025core,chen2025customcontrast}. From a handful of references, a method must bind a concept to a textual handle, recover its fine-grained appearance, and follow new contexts, attributes, and interactions. The evidence should also be selectively addressable: inactive when its handle is absent and composable without identity omission or attribute leakage \citep{huang2025multicondition,yao2025conceptconductor,peng2026tara}. Figure~\ref{fig:front-showcase} illustrates this objective for one toy car across different environments and actions.

Existing methods expose a capacity--modularity trade-off. Textual Inversion stores a concept in a small token embedding but often lacks capacity for detailed identity \citep{gal2023textual,wu2025core}. DreamBooth and LoRA increase capacity by adapting generator pathways, yielding concept-coupled state whose composition can introduce interference \citep{ruiz2023dreambooth,hu2022lora,peng2026tara}. Encoder-based systems instead encode references at inference time for instant general-subject or face conditioning \citep{ye2023ipadapter,zhang2024ssrencoder,li2024photomaker,wang2024instantid}. Together, these methods leave room for memory richer than a single token embedding but still explicitly addressed by text and inactive outside that address. This is particularly relevant to \sdthree{}, where text and image tokens interact throughout \mmdit{} blocks rather than through a fixed U-Net cross-attention interface \citep{esser2024scaling,peebles2023dit,feng2026personalize}.

We propose \method{}, a trigger-indexed external memory for \sdthree{}. Each concept receives an explicit trigger such as \sampletrigger{}, which registers tokenizer-specific $n$-gram entries. Exact matches activate the entries present in a prompt; their directions and tanh gates enter mounted text-encoder and \mmdit{} context states as residuals scaled to the local hidden-state norm. Concept evidence resides outside the backbone, and all pretrained components remain frozen.

Unmatched prompts receive no Engram residual; prompts containing registered triggers activate and sum all exactly matched entries. The interface can acquire a novel subject, separate similar instances, or refine an imprecise familiar concept. Storage grows with the registry, while inference activates only prompt-matched entries.

Experiments compare \method{} with zero-shot \sdthree{}, embedding-only learning, and joint DreamBooth-LoRA. \method{} and LoRA show broadly comparable fidelity with complementary CLIP-I and DINOv2 outcomes. In held-out contexts, \method{} preserves identity better, while LoRA favors prompt alignment. Qualitative studies show joint expression of separately addressed memories, inactive-route preservation, and same-class separation. Ablations support gated residual injection and distributed \mmdit{} access, while showing that denser mounting is not uniformly beneficial. Overall, \method{} offers a different balance of fidelity, modularity, and compositional access rather than a universally stronger replacement for weight adaptation.

Our contributions are:
\begin{itemize}
    \item We formulate few-shot acquisition and refinement as exact trigger-indexed memory retrieval over a frozen diffusion transformer.
    \item We introduce concept-specific Engram directions and gated relative residual injection for explicit, prompt-selective memory access in SD3.5.
    \item We characterize its trade-off against embedding-only learning and joint LoRA across fidelity, addressable loading, and compositional access.
\end{itemize}

\begin{figure}[t]
\centering
\includegraphics[width=\textwidth]{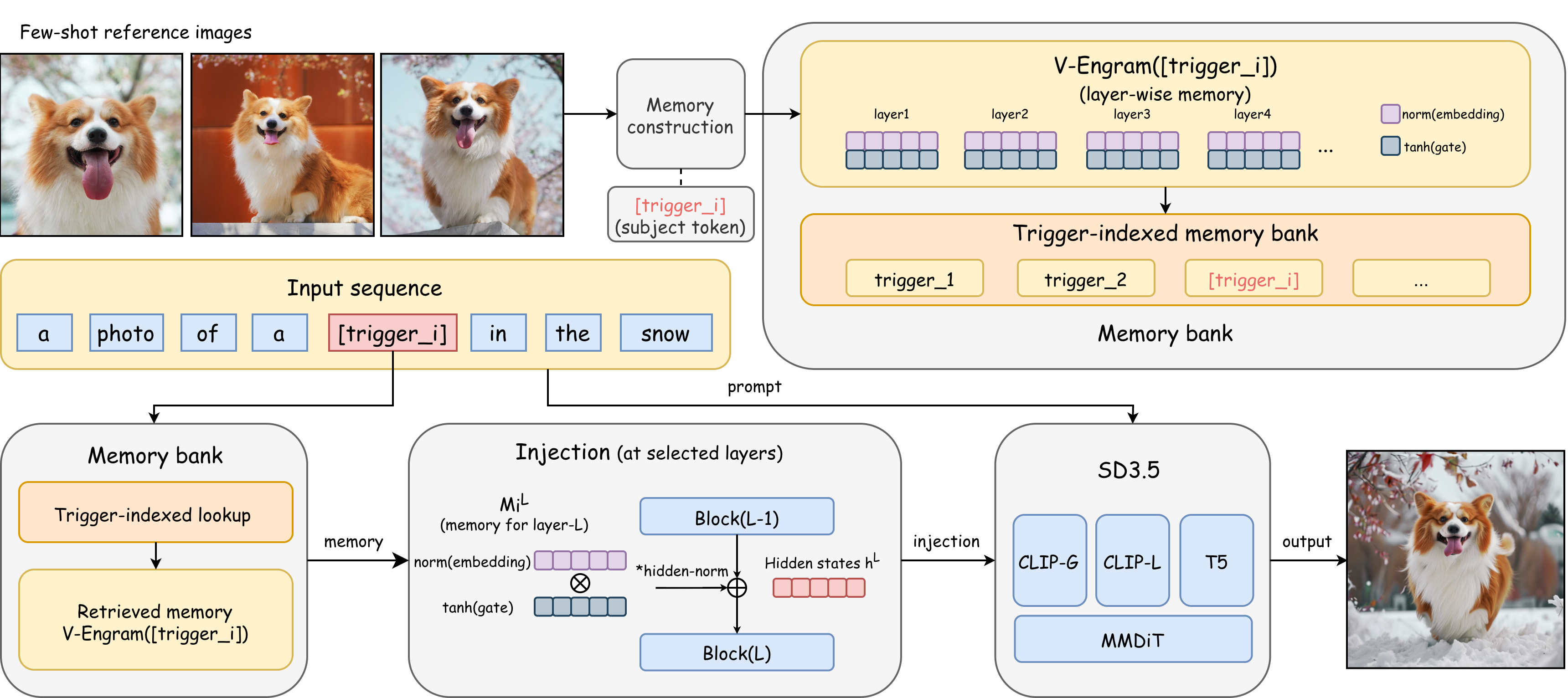}
\caption{Overview of \method{}. Few-shot references and an explicit trigger construct tokenizer-specific, layer-wise Engram entries. Exact token-id matches retrieve entries whose residuals are injected into selected frozen \sdthree{} layers.}
\label{fig:tinyengram-overview}
\end{figure}

\section{Related Work}

\paragraph{Conditional Memory.}
Parameter-efficient adaptation limits trainable state through adapters, prompts, or low-rank updates, but does not by itself specify concept-level activation \citep{houlsby2019parameter,li2021prefix,lester2021power,hu2022lora,liu2022fewshot}. External memory adds an addressing mechanism: Product-Key Memory performs structured content lookup, Memorizing Transformers use approximate $k$NN retrieval, and DeepSeek Engram deterministically maps matched token $n$-grams to learned entries \citep{lample2019productkey,wu2022memorizing,cheng2026conditional}. \method{} shares deterministic lexical addressing, but changes both the memory content and target: few-shot visual references train concept-specific residual directions and gates for a frozen diffusion model, without semantic or approximate retrieval.

\paragraph{Optimization-Based Personalization.}
Learned concepts can reside in token space or model updates. Textual Inversion learns one embedding, CoRe regularizes its contextual interaction, and NeTI and ProSpect enrich frozen-backbone representations across layers or timesteps \citep{gal2023textual,wu2025core,alaluf2023neti,zhang2023prospect}. DreamBooth fine-tunes with prior preservation, while compact alternatives update attention, low-rank or rank-one factors, singular values, or sparse concept neurons \citep{ruiz2023dreambooth,kumari2023custom,hu2022lora,tewel2023key,han2023svdiff,liu2023cones}. CustomContrast and AttnDreamBooth further target subject discrimination or text alignment \citep{chen2025customcontrast,pang2024attndreambooth}. \method{} occupies a different point: the backbone remains frozen as in embedding learning, but its external, layer-wise memory is not confined to one input embedding.

\paragraph{Reference-Guided Methods.}
Image references can produce conditioning or compact adaptation state. BLIP-Diffusion, ELITE, IP-Adapter, and SSR-Encoder learn general subject features \citep{li2023blipdiffusion,wei2023elite,ye2023ipadapter,zhang2024ssrencoder}; PhotoMaker and InstantID specialize in human identity \citep{li2024photomaker,wang2024instantid}; and HyperDreamBooth and AnyDoor predict compact weights or object features \citep{ruiz2023hyperdreambooth,chen2023anydoor}. These methods make a reference encoder or reference image part of the personalization interface. \method{} instead distills references once into persistent, text-addressed memory, so generation is subsequently invoked by the trigger alone.

\paragraph{Multi-Concept Methods.}
Prior work jointly learns concepts, extracts several handles, or fuses adapters \citep{kumari2023custom,avrahami2023break,gu2023mixofshow}. FastComposer and DynASyn target identity and action control, while condition merging and isolated sampling paths reduce multi-condition confusion and leakage \citep{xiao2023fastcomposer,choi2025dynasyn,huang2025multicondition,yao2025conceptconductor}. TARA reduces interference in LoRA composition through rare-token masking and token--region alignment \citep{peng2026tara}. \method{} instead stores trigger-indexed direction-and-gate entries and injects them as exact-match hidden-state residuals, shifting the focus from spatially aligned LoRA composition to deterministic retrieval and selective activation of persistent concept memory.

\paragraph{Diffusion Transformers and \mmdit{}.}
Latent diffusion commonly uses a text-conditioned U-Net, whereas DiT replaces it with transformer blocks \citep{rombach2022latent,peebles2023dit}. \sdthree{} combines CLIP/OpenCLIP and T5 streams in \mmdit{}, where text and image tokens interact jointly \citep{radford2021clip,ilharco2021openclip,raffel2020t5,esser2024scaling}. Recent work uses DiT features for training-free subject injection, but MMDiT can still neglect or mix similar subjects \citep{feng2026personalize,wei2024enmddit}. \method{} therefore targets concept-specific text and context states inside SD3.5, beyond an initial embedding or U-Net cross-attention interface.

\section{Method}

\subsection{Overview}

Figure~\ref{fig:tinyengram-overview} shows the workflow. A few references train external entries registered from a trigger such as \sampletrigger{}. At inference, exact tokenizer-specific span matches retrieve these entries; unmatched prompts follow the frozen path. Retrieved entries enter selected hidden-state sites as gated residuals.

Let $G_\theta$ be a pretrained \sdthree{} pipeline with frozen text encoders, VAE, and \mmdit{} transformer. A concept $c$ is specified by a few reference images $\mathcal{D}_c=\{x_i^c\}_{i=1}^{N_c}$ and a trigger string $s_c$ that appears in training prompts, e.g., ``a photo of \sampletrigger{}''. The trainable concept memory $\phi_c$ consists of Engram directions and tanh gates attached to a set of mounted sites $\mathcal{Q}$. Each site $q=(g,\ell)$ denotes a tokenizer/conditioning stream $g$ and a hidden-state location $\ell$, with hidden width $d_q$; we write $g(q)$ for the stream associated with site $q$. The backbone parameters $\theta$ are never updated.

\subsection{Trigger-Indexed Memory Registry}

\sdthree{} uses three tokenizer streams, CLIP-L, OpenCLIP-G, and T5-XXL; denote this set by $\mathcal{G}$. Since the same trigger can tokenize differently across streams, \method{} builds a separate registry for each tokenizer. The registry for stream $g\in\mathcal{G}$ contains token sequences and the memory entries they address. For a prompt $p$, exact token-id scanning returns
\begin{equation}
    \mathcal{A}_g(p)=
    \{(m,i): \tau_g(p)_{i:i+|\kappa_m|}=\kappa_m,\ \kappa_m\in\mathcal{R}_g\},
\end{equation}
where $\kappa_m$ is a registered span, $m$ its Engram entry, and $i$ its start position. The same surface trigger may tokenize differently across streams. The main configuration registers contiguous 2--4-grams and always retains the full trigger span; single-token triggers use exact spans. All exact matches, including overlaps, are active and summed below. This decomposition is intended to preserve local relations among constituent tokens in a multi-token trigger, a property not separately evaluated here. Lookup uses token ids rather than semantic or approximate retrieval, attention-based detection, or prompt rewriting. With no match, the conditioning path equals the frozen base model.

\subsection{Gated Relative Residual Injection}

For each registered entry $m$ and mounted site $q=(g,\ell)$, \method{} stores a trainable direction $e_{m,q}\in\mathbb{R}^{d_q}$ and a per-channel gate $a_{m,q}\in\mathbb{R}^{d_q}$. Routing is determined by the trigger lookup; the gate modulates the signed residual strength and is not a separate routing network. Since naive additive memories can over-perturb frozen hidden states, especially when mounted at several transformer blocks, \method{} injects a gated relative residual. Given the hidden-state sequence $H^q$ of length $T_q$ at site $q$, the residual produced by entry $m$ is
\begin{equation}
\begin{aligned}
    \Delta_{m,q}(H^q)
    &=
    \rho(H^q)
    \left(
    \frac{e_{m,q}}{\|e_{m,q}\|_2+\varepsilon}
    \odot
    \tanh(a_{m,q})
    \right), \\
    \rho(H^q)
    &=
    \operatorname{sg}
    \left(\frac{1}{T_q}\sum\nolimits_{j=1}^{T_q} \|H^q_j\|_2\right).
\end{aligned}
\end{equation}
The Engram direction supplies residual direction, the tanh gate provides bounded signed per-channel modulation, and the detached norm $\rho(H^q)$ calibrates the current activation scale. Our single-vector direct-embedding baseline disables this entire path and injects the learned embedding without a gate, direction normalization, or relative scaling:
\begin{equation}
    \Delta^{\mathrm{ungated}}_{m,q}
    =
    e_{m,q}.
\end{equation}
For prompt $p$, let $\mathcal{C}^{q}_{j}(p)$ collect the matched entries whose trigger spans cover position $j$ at site $q$. An entry $m$ belongs to $\mathcal{C}^{q}_{j}(p)$ if its matched span in $\mathcal{A}_{g(q)}(p)$ covers token position $j$. Omitting the repeated argument of $\Delta_{m,q}(H^q)$, the update is
\begin{equation}
    \tilde{H}^{q}_{j}
    =
    H^q_j+
    \sum\nolimits_{m\in\mathcal{C}^{q}_{j}(p)}
    \Delta_{m,q}.
\end{equation}
This form covers both the full trigger span and the case where a trigger such as \sampletrigger{} activates several registered $n$-gram entries over the same token region. For CLIP-L and OpenCLIP-G, the injection is mounted before the penultimate transformer block, so downstream CLIP states used by \sdthree{} are Engram-conditioned. For T5-XXL, it is mounted before the final encoder block and affects the final T5 hidden states.

\subsection{Memory Mounting in \sdthree{}}

\method{} first attaches memory to the three text-encoder streams and also mounts memory inside selected \mmdit{} text/context blocks. We denote the selected MMDiT blocks by $\mathcal{L}_{\mathrm{mmdit}}$. Unless otherwise specified, our default configuration uses seven evenly distributed blocks, $\mathcal{L}_{\mathrm{mmdit}}=\{0,6,12,19,25,31,37\}$. Encoder-only mounting and different \mmdit{} mounting densities are evaluated in the ablation study. After SD3.5 concatenates CLIP and T5 contexts, matched CLIP spans map to the CLIP context region and matched T5 spans map to the T5 region. We maintain tokenizer-specific offsets when constructing this concatenated context sequence, so each trigger match is injected only into the corresponding CLIP or T5 token region. The pooled projection path is left unchanged. This gives a simple way to vary how deeply concept memory enters the multimodal denoising computation without modifying the pretrained transformer weights.

\subsection{Training Objective and Storage}

\method{} stores external memory outside the pretrained model as small state dictionaries: three text-encoder Engram banks and their gate banks, plus selected \mmdit{} text/context banks. The trigger configuration records the trigger strings, tokenizer-specific registered targets, mounting sites, and gate setting. At inference time, the registry identifies exact prompt-matched entries, and only these entries are activated. Multiple concepts are composed by matching multiple triggers and summing all matched residuals; no LoRA adapters are merged and no backbone weights are modified.

For a concept $c$, let $K_{c,g}=|\mathcal{M}_{c,g}|$ be the number of entries registered in stream $g$. The trainable storage scales with registered entries and mounted sites:
\begin{equation}
    |\phi_c|_{\mathrm{params}}
    =
    2\sum\nolimits_{q\in\mathcal{Q}} K_{c,g(q)}d_q,
\end{equation}
where $\mathcal{M}_{c,g}$ is the set of entries registered for concept $c$ in stream $g$ and the factor $2$ corresponds to one direction vector and one gate vector. The ungated direct-embedding variant stores only the embedding vectors, giving $\sum\nolimits_{q\in\mathcal{Q}} K_{c,g(q)}d_q$ parameters.

Training freezes all pretrained components and updates only Engram directions and gates. For a clean latent $z_0$, Gaussian noise $\epsilon$, and flow time $t$, define $z_t=(1-t)z_0+t\epsilon$ and the SD3.5 rectified-flow target $v=\epsilon-z_0$. The objective is
\begin{equation}
    \mathcal{L}_{\mathrm{flow}} =
    \mathbb{E}_{z_0,\epsilon,t,p}
    \left[
    \left\|F_\theta(z_t,t,p;\phi_{\mathcal{A}(p)}) - v \right\|_2^2
    \right],
\end{equation}
where $\mathcal{A}(p)=\bigcup_{g\in\mathcal{G}}\mathcal{A}_g(p)$ denotes the active prompt-matched Engram entries across the three tokenizer streams. Prompts without active triggers use the unmodified frozen conditioning path.

\section{Experiments}

\subsection{Experimental Setup}

\paragraph{Benchmarks and references.}
The main benchmark contains 15 subjects and 80 Google DreamBooth references \citep{ruiz2023dreambooth}. Same-class disambiguation uses seven dogs and 37 references from the same dataset. For familiar-identity refinement, we select 15 Celebrity-1000 identities \citep{tonyassi2026celebrity1000} and form, for each identity, a five-image training set and a five-image-disjoint test set with visibly varied pose, hairstyle, illumination, and setting.

\paragraph{Baselines and training.}
We compare frozen \sdthree{}, TI-style embedding learning \citep{gal2023textual}, joint DreamBooth-LoRA, and \method{}. All learned methods use the same references and joint 15-subject training. TI-style, both LoRA ranks, and \method{} use 25k, 8k, and 20k steps at learning rates $10^{-3}$, $10^{-4}$, and $5\times10^{-4}$, respectively. All use batch size 4, AdamW, a constant schedule, bfloat16, and seed 42. By default, \method{} uses gated memories in the text encoders and seven \mmdit{} blocks $\{0,6,12,19,25,31,37\}$. Rank 4 (5.895M parameters) is the fixed LoRA configuration for controlled analyses; rank 8 (11.790M) is additionally evaluated in the main and contextual comparisons as a capacity check near the 10.477M full \method{} registry. All qualitative figures containing LoRA outputs use rank 4.

\paragraph{Inference and metrics.}
To isolate subject fidelity, the main comparison uses ``a photo of \texttt{<trigger>}'', with the subject-specific address substituted for \texttt{<trigger>}. Seeds 42--46 yield five $768\times768$ images per subject using 20 steps, guidance 4.5, the default scheduler, no negative prompt, and bfloat16. CLIP-I and DINOv2 use CLIP-L/14 and DINOv2-base cosine similarity \citep{radford2021clip,oquab2024dinov2}. Best-reference takes each generation's maximum over its references; all-pairs averages the complete generation--reference matrix. Both are averaged within subject and then macro-averaged, respectively tolerating viewpoint variation and measuring full-reference consistency. CLIP-T encodes the exact contextual prompt, including its angle-bracketed trigger. For familiar identities, we compute cosine similarity between ArcFace-style face embeddings \citep{deng2019arcface} extracted with InsightFace's \texttt{buffalo\_l} model pack, comparing generations only against held-out references and reporting valid-face coverage separately. Further details are supplementary.

\subsection{Main Comparison}

Table~\ref{tab:main-comparison} asks whether trigger-indexed memory can match weight adaptation while improving over input-level embedding learning.

\begin{table}[!ht]
\centering
\small
\setlength{\tabcolsep}{2.7pt}
\begin{tabular}{@{}lrrrr@{}}
\toprule
& \multicolumn{2}{c}{Avg. best-ref} & \multicolumn{2}{c}{Avg. all-pairs} \\
\cmidrule(lr){2-3}\cmidrule(l){4-5}
Method & DINOv2 $\uparrow$ & CLIP-I $\uparrow$ & DINOv2 $\uparrow$ & CLIP-I $\uparrow$ \\
\midrule
Zero-shot SD3.5 & 0.4274 & 0.6765 & 0.3501 & 0.6361 \\
TI-style & 0.4919 & 0.7188 & 0.4071 & 0.6813 \\
DB-LoRA ($r=4$) & 0.8084 & 0.8682 & 0.6911 & 0.8290 \\
DB-LoRA ($r=8$) & \textbf{0.8167} & 0.8875 & \textbf{0.6930} & 0.8439 \\
\method{} & 0.7896 & \textbf{0.8902} & 0.6809 & \textbf{0.8463} \\
\bottomrule
\end{tabular}
\caption{Main comparison over 15 subjects. Both reference aggregations are averaged over five generations and then macro-averaged over subjects.}
\label{tab:main-comparison}
\end{table}

\begin{figure}[!ht]
    \centering
    \includegraphics[width=\resultfigurewidth]{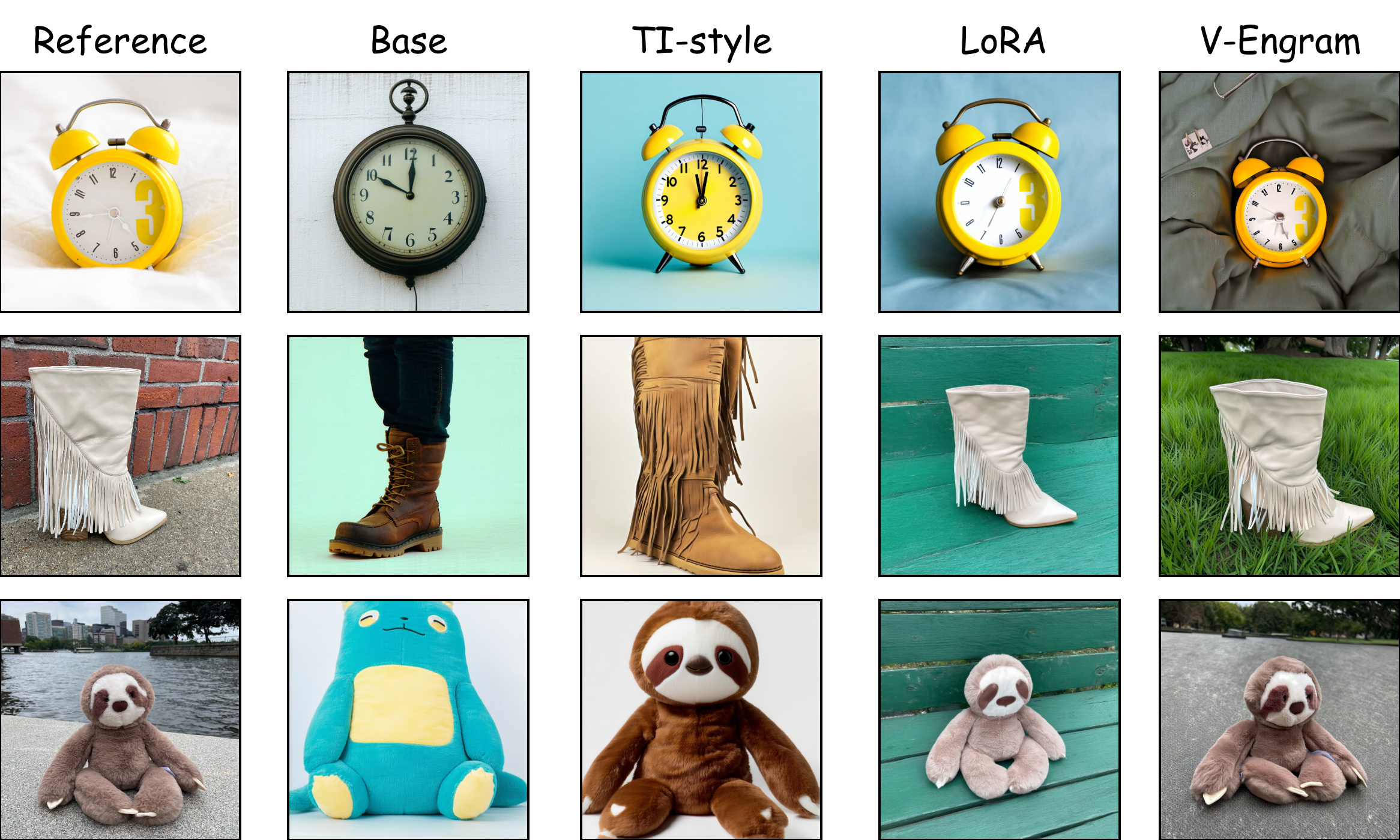}
    \caption{Qualitative main comparison under the common prompt template ``a photo of \texttt{<trigger>}''. Rows show the personalized clock, fancy boot, and gray sloth plushie; columns show one reference image, zero-shot SD3.5, TI-style / embedding-only personalization, rank-4 DreamBooth-LoRA, and \method{}.}
    \label{fig:clock-main}
\end{figure}

Figure~\ref{fig:clock-main} illustrates the rank-4 qualitative comparison for three representative subjects.

\FloatBarrier

The base and TI-style outputs frequently drift at the instance level, whereas rank-4 LoRA and \method{} more faithfully recover distinguishing appearance, including the clock's twin-bell silhouette and yellow numeral motif, the boot's fringe structure, and the gray sloth plushie's shape and facial details. Increasing LoRA from rank 4 to 8 raises best-reference CLIP-I by 0.0192 and DINOv2 by 0.0082. Against the parameter-near rank-8 adapter, \method{} is within 0.0028 on both CLIP-I aggregations, while rank-8 LoRA leads DINOv2 by 0.0271 best-reference and 0.0121 all-pairs. Taken together, the two methods achieve broadly comparable subject fidelity under the simple prompt: \method{} essentially matches rank-8 LoRA in CLIP-I, whereas LoRA is higher in DINOv2, so neither dominates across visual representations.

\paragraph{Resource scaling.}
We measure addressability as the registry grows. Joint rank-4 LoRA remains fixed at 5.895M parameters; nested \method{} registries with 1, 7, and 9 triggers contain 0.639M, 4.985M, and 6.221M, crossing LoRA at nine. In a controlled A40 profile (10 warmup and 100 measured steps), \method{} uses 0.940--1.004 GiB less peak allocated training memory but takes 14.9--29.0\% longer per step. For a single-trigger query to a 15-trigger checkpoint, matched loading activates 1.061M rather than 10.477M Engram parameters, avoiding 89.87\% of full-registry adaptation-state loading. This enables selective adaptation-state management without loading unrelated memories. Per-setting measurements are supplementary.

\subsection{Ablation Study}

We isolate injection depth and the gated residual. The density study jointly trains five concepts (28 references); all variants use 10k steps, five seeds per concept, and fixed settings. Encoder-only retains three text-encoder sites; the others additionally use $N=3$--9 uniformly distributed \mmdit{} blocks.

\begin{table}[!ht]
\centering
\small
\setlength{\tabcolsep}{7pt}
\begin{tabular}{@{}lrr@{}}
\toprule
Mounting setting & DINOv2 $\uparrow$ & CLIP-I $\uparrow$ \\
\midrule
Encoder only & 0.6132 & 0.7991 \\
$N=3$ & 0.7612 & 0.8499 \\
$N=4$ & 0.7814 & 0.8719 \\
$N=5$ & 0.7595 & 0.8820 \\
$N=6$ & 0.7780 & 0.8894 \\
$N=7$ & \textbf{0.8068} & \textbf{0.9007} \\
$N=8$ & 0.7345 & 0.8692 \\
$N=9$ & 0.8058 & 0.8798 \\
\bottomrule
\end{tabular}
\caption{Mounting-density ablation. $N$ is the number of \mmdit{} blocks in addition to encoder sites.}
\label{tab:ablations}
\end{table}

\begin{figure}[!ht]
    \centering
    \includegraphics[width=\resultfigurewidth]{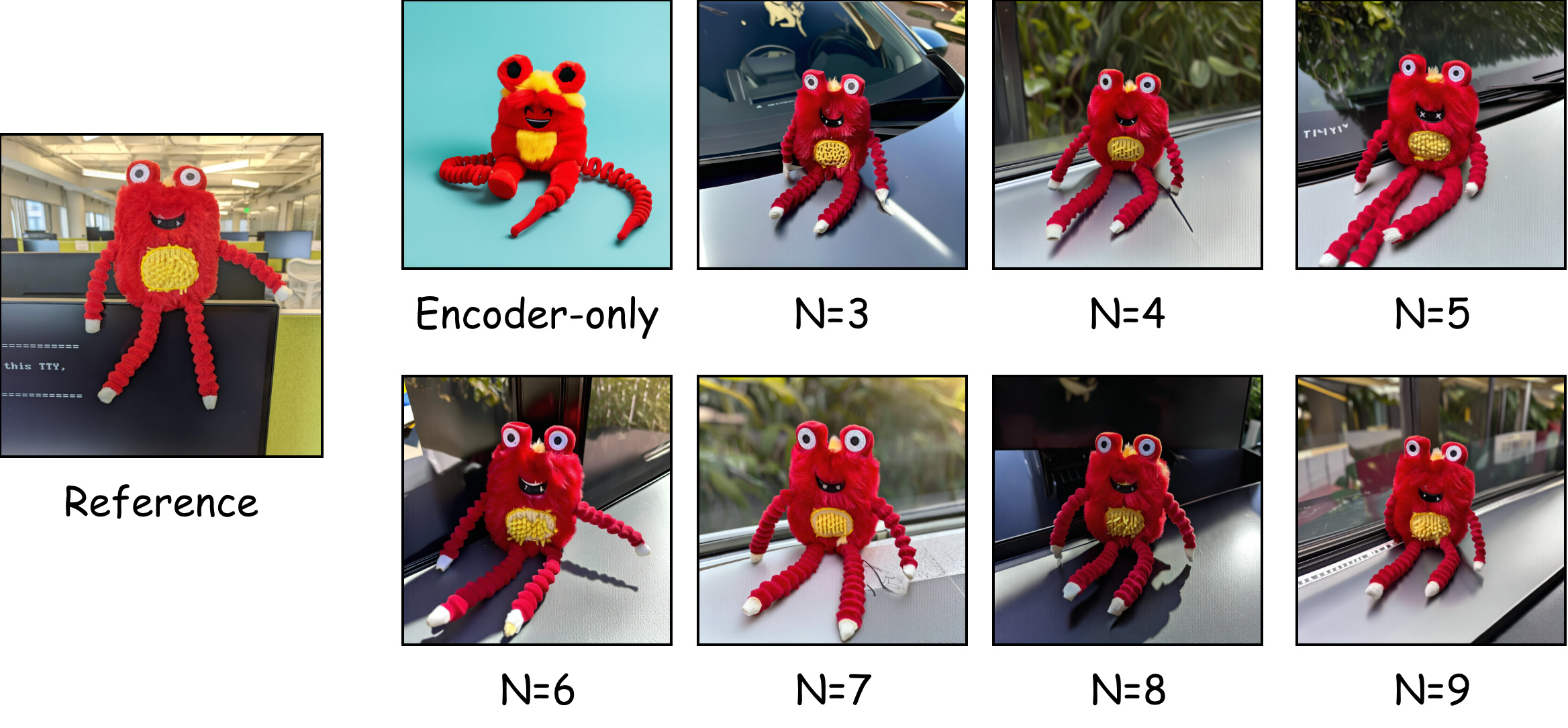}
    \caption{Qualitative mounting-density ablation. $N$ denotes the number of selected \mmdit{} blocks. Compared with encoder-only injection, mounted variants more consistently preserve the subject's distinctive shape and appearance, with visible variation across densities.}
    \label{fig:ablation-qual}
\end{figure}

Figure~\ref{fig:ablation-qual} shows the corresponding qualitative variation.

\FloatBarrier

All mounted variants beat encoder-only on both metrics, while density is non-monotonic. Seven blocks perform best on this validation subset and define our default, without implying an architecture-independent optimum. A matched rerun compares TI-like direct single-vector injection with the complete V-Engram residual---normalized directions, hidden-state-relative scaling, and per-channel gates---under fixed data, schedule, and sites. The full design raises DINOv2 from 0.7065 to 0.8113 and CLIP-I from 0.8262 to 0.8892.

\subsection{Additional Analysis}

We next test held-out contexts, paired memories, unmatched routing, same-class instances, and refinement of imprecise prior knowledge.

\paragraph{Contextual Composition.}
We apply five held-out, category-agnostic templates---close-up, room, snow, garden, and beside a red car---to all 15 subjects at seed 42. Shared wording enables controlled comparison across objects and animals without category-specific prompt engineering. Reusing the main checkpoints gives 75 images per method; Figure~\ref{fig:front-showcase} separately shows broader composition for one toy car.

\begin{table}[!ht]
\centering
\small
\setlength{\tabcolsep}{5pt}
\begin{tabular}{@{}lrrr@{}}
\toprule
& CLIP-T & \multicolumn{2}{c}{CLIP-I} \\
\cmidrule(lr){2-2}\cmidrule(l){3-4}
Method & Score $\uparrow$ & Best-ref $\uparrow$ & All-pairs $\uparrow$ \\
\midrule
DB-LoRA ($r=4$) & 0.2558 & 0.8264 & 0.7888 \\
DB-LoRA ($r=8$) & \textbf{0.2658} & 0.8105 & 0.7706 \\
\method{} & 0.2334 & \textbf{0.8717} & \textbf{0.8261} \\
\bottomrule
\end{tabular}
\caption{Contextual composition under five category-agnostic templates shared by all 15 subjects. Scores are macro-averaged over subjects.}
\label{tab:contextual-composition}
\end{table}

\begin{figure}[!ht]
    \centering
    \includegraphics[width=\resultfigurewidth]{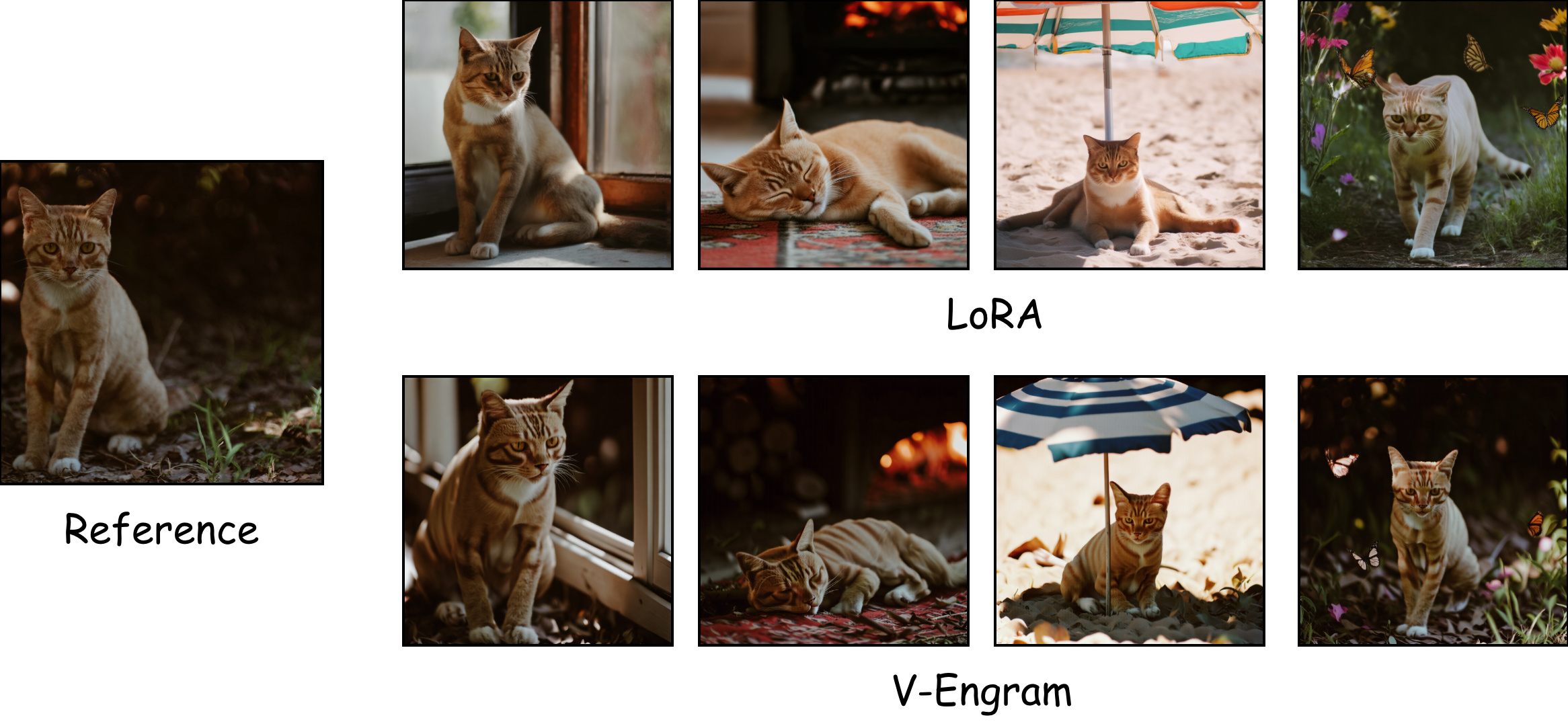}
    \caption{Additional contextual prompts for a personalized cat, shown separately from the five-template protocol in Table~\ref{tab:contextual-composition}. The reference is at left; the upper and lower rows show rank-4 DreamBooth-LoRA and \method{}, respectively. Exact prompts and additional examples are provided in the supplementary material.}
    \label{fig:contextual-composition}
\end{figure}

Table~\ref{tab:contextual-composition} shows that \method{} better preserves personalized subjects in complex contexts, whereas rank-8 LoRA aligns text better. \method{} leads rank 8 by 0.0612/0.0554 in best-reference/all-pairs CLIP-I (14/15 subjects), while rank 8 leads CLIP-T by 0.0324 (13/15). Raising LoRA from rank 4 to 8 improves CLIP-T by 0.0101 but reduces the two CLIP-I scores by 0.0158/0.0182; more capacity thus does not improve contextual fidelity. Figure~\ref{fig:contextual-composition} illustrates this trade-off.

\begin{figure}[!ht]
    \centering
    \includegraphics[width=\resultfigurewidth]{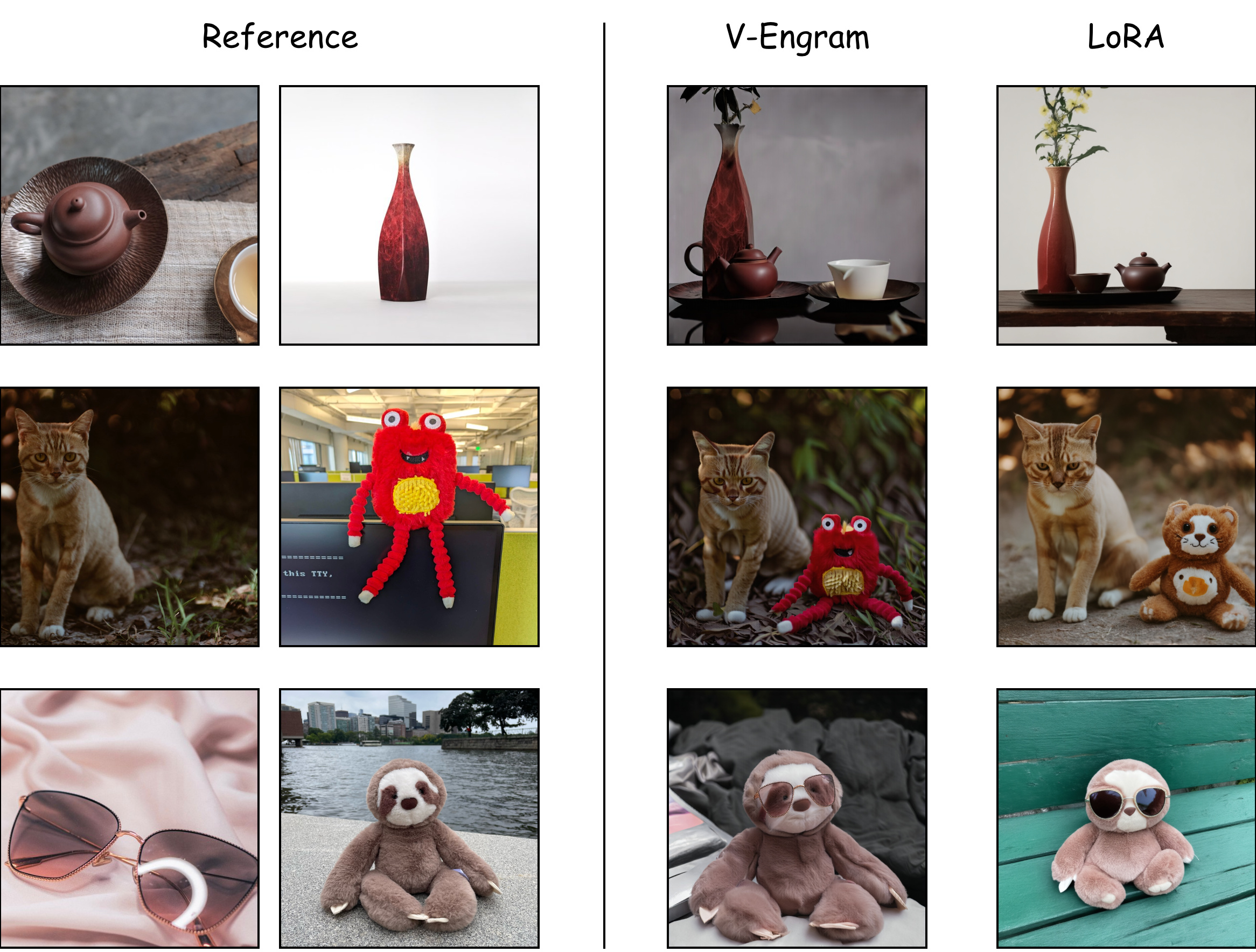}
    \caption{Paired-trigger examples: teapot+vase, cat+toy, and sunglasses+plushie. Columns show two references, \method{}, and rank-4 DreamBooth-LoRA.}
    \label{fig:multi-concept-qual}
\end{figure}

\paragraph{Multi-Concept Composition.}
Figure~\ref{fig:multi-concept-qual} shows three of 20 matched seed-42 paired-trigger prompts against rank-4 LoRA. \method{} more consistently depicts both concepts, while LoRA sometimes omits or blends one. We keep this comparison qualitative because full-image similarity cannot attribute retention to each subject; both methods still exhibit failures.

\FloatBarrier

\paragraph{Non-Triggered Base-Behavior Preservation.}
Eight ordinary categories under seven templates yield 56 prompts without registered triggers. Exact-match logs show zero active entries in all three text streams; no residual is injected, so \method{} exactly reduces to the frozen base under identical inference settings. Joint LoRA remains globally active and can shift outputs toward training instances. Supplementary examples visualize this difference.

\begin{figure}[!t]
    \centering
    \includegraphics[width=\resultfigurewidth]{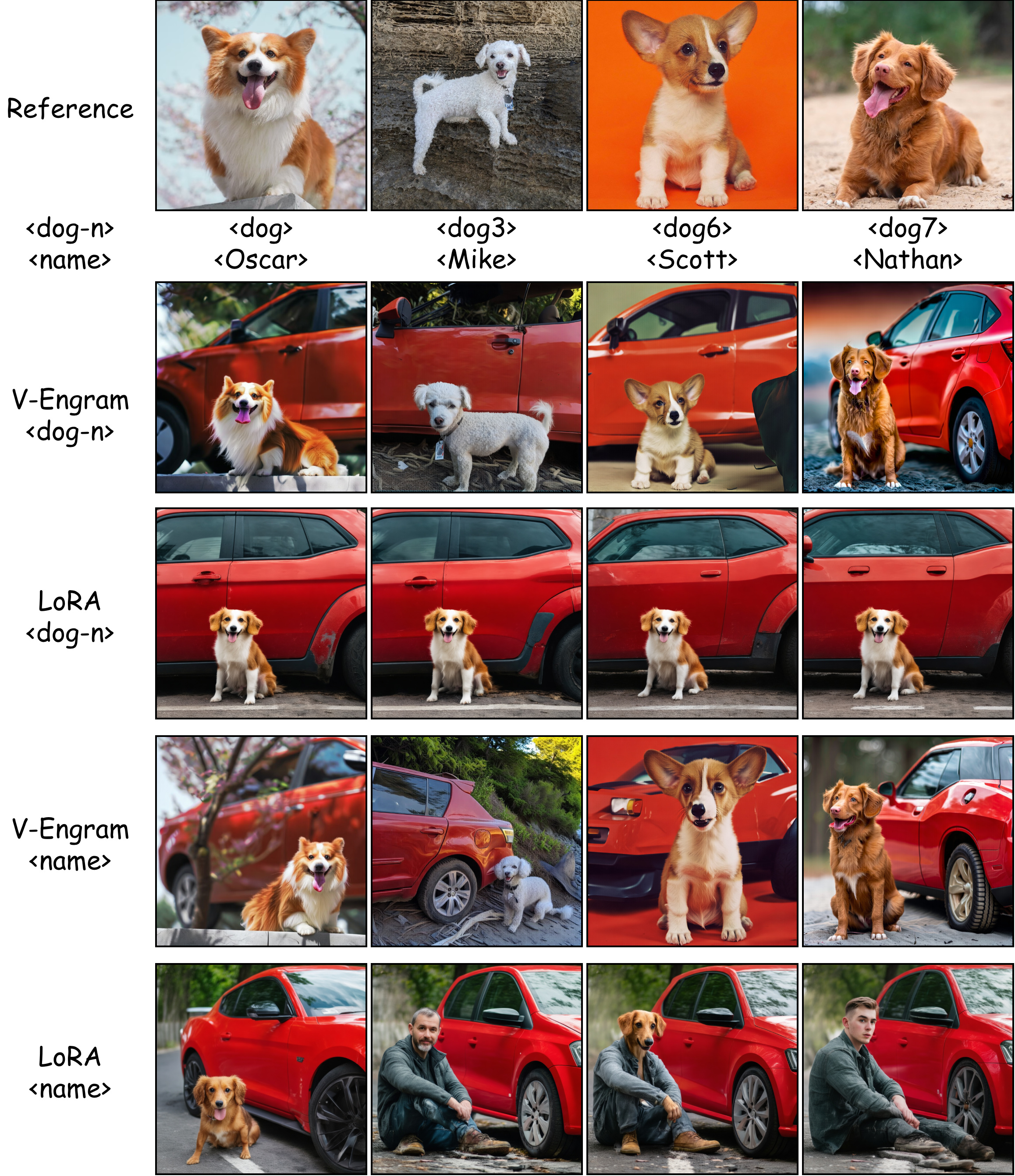}
    \caption{Same-class disambiguation under indexed and name-based triggers for the red-car prompt. Four of the seven dog identities are shown; rows compare \method{} and rank-4 DreamBooth-LoRA under the two naming schemes.}
    \label{fig:same-class-dogs}
\end{figure}

\paragraph{Same-Class Trigger Disambiguation.}
Seven dogs (37 references) receive indexed triggers (\texttt{\textless dog\textgreater}, \texttt{\textless dog2\textgreater}, \ldots) or arbitrary names (\texttt{\textless Oscar\textgreater}, \texttt{\textless Ryan\textgreater}, \ldots). \method{} trains for 10k steps and rank-4 LoRA for 4k. Seed-42 results cover a simple photo, running on grass, and sitting beside a red car; Figure~\ref{fig:same-class-dogs} shows the red-car slice for four identities. Because its exact-match registry accepts user-defined triggers, \method{} is not tied to one naming convention; indexed and name-based triggers are the two forms tested here, and both better preserve dog-specific cues. LoRA tends toward similar dogs with indexed triggers and can invoke human priors with names. Without a calibrated dog-identity recognizer, this remains qualitative.

\paragraph{Refining Familiar Concepts.}
Finally, we test calibration of imprecise prior knowledge on 15 identities. Each condition generates five images per identity, compared only with five held-out references. The image-disjoint splits contain photographs from different times and capture conditions, rewarding recurring identity cues rather than training-image reconstruction.

\begin{table}[!ht]
\centering
\small
\setlength{\tabcolsep}{4pt}
\begin{tabular}{@{}lrrr@{}}
\toprule
& \multicolumn{2}{c}{InsightFace} & \\
\cmidrule(lr){2-3}
Method & Best-ref $\uparrow$ & All-pairs $\uparrow$ & Valid \\
\midrule
Zero-shot \sdthree{} & 0.2344 & 0.1824 & 75/75 \\
DB-LoRA ($r=4$) & 0.4618 & 0.4014 & 75/75 \\
\method{} & \textbf{0.6084} & \textbf{0.5415} & 75/75 \\
\bottomrule
\end{tabular}
\caption{InsightFace identity similarity on 15 familiar identities. Scores are averaged over generations and macro-averaged over identities.}
\label{tab:familiar-identity}
\end{table}

\begin{figure}[!ht]
    \centering
    \includegraphics[width=\resultfigurewidth]{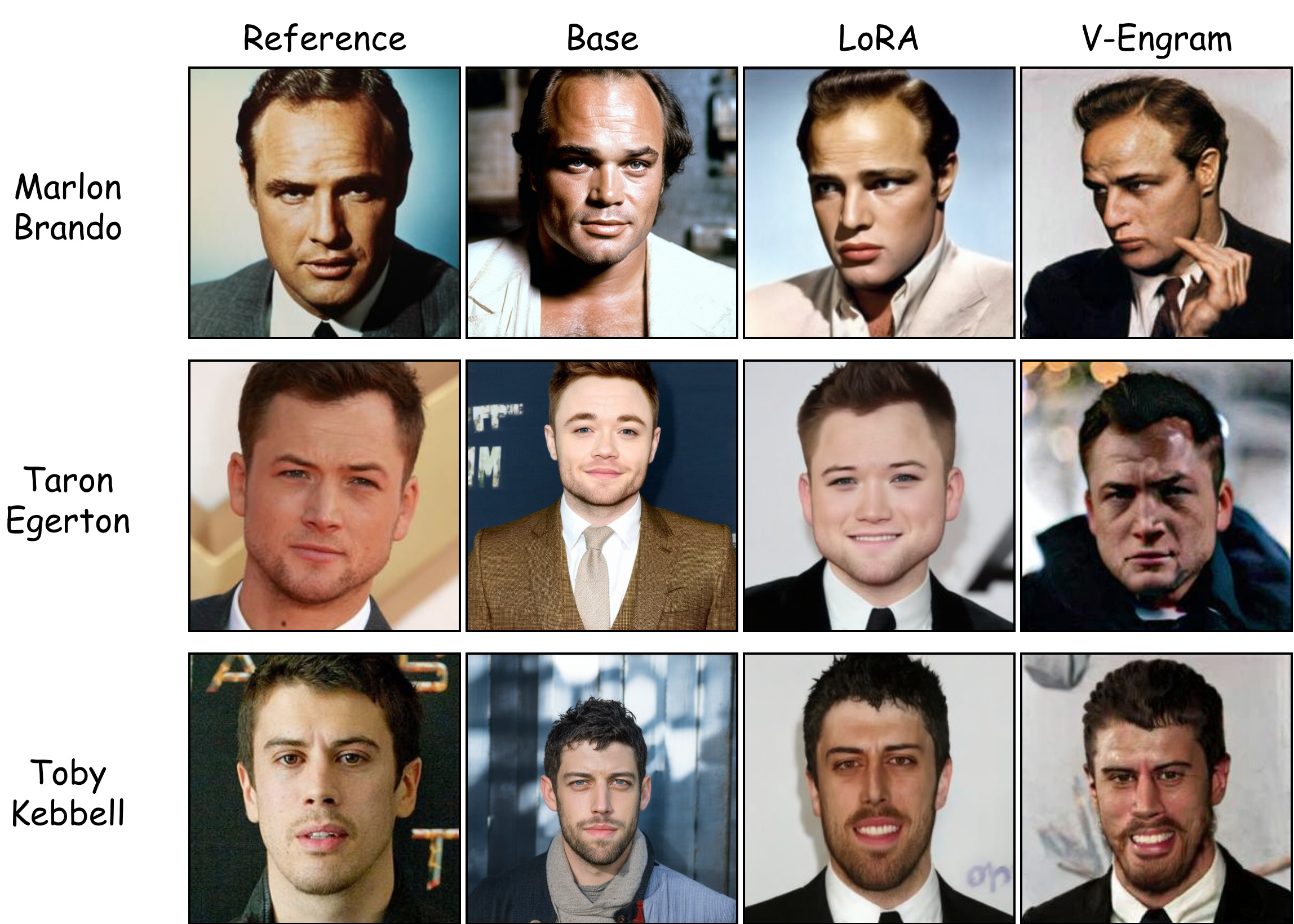}
    \caption{Qualitative familiar-identity refinement for three of the 15 evaluated identities. Columns show a held-out reference, zero-shot \sdthree{}, rank-4 DreamBooth-LoRA, and \method{}.}
    \label{fig:familiar-identity}
\end{figure}

\method{} exceeds the tested rank-4 LoRA by 0.1466 best-reference and 0.1401 all-pairs similarity, winning 13/15 identities; both adaptations beat the base on all 15. All 75 generations per method are valid detections. The image-disjoint split supports identity-level refinement beyond single-image memorization, with three held-out visual examples shown in Figure~\ref{fig:familiar-identity}.

\FloatBarrier

\section{Conclusion}

We introduced \method{}, a trigger-indexed external-memory interface that provides a frozen \sdthree{} backbone with prompt-addressed Engram entries. Under simple subject prompts, \method{} is broadly competitive with LoRA, with comparable CLIP-I but lower DINOv2; in held-out contexts, it preserves identity better, while LoRA favors prompt alignment. Exact $n$-gram routing leaves unmatched prompts on the frozen base path and supports selective memory activation and loading; further analyses support familiar-identity refinement, paired-trigger composition, and same-class separation. Overall, these findings show a different balance among fidelity, modularity, and compositional access rather than universal superiority over weight adaptation. More broadly, exact token matching may provide an addressing mechanism for external memory in other token-conditioned generative or multimodal architectures, although cross-architecture validation remains future work.

\bibliographystyle{plainnat}
\bibliography{refs}

\end{document}